# Measuring AI Leadership: Development and Validation of a Multidimensional Measure for AI-Native Organizations

Mustafa Akben[1] and Leslie Coyne[2]

[1] Department of Management and Entrepreneurship

Love School of Business, Elon University

[2] Signalead LLC

## Author Note

Mustafa Akben, PhD, Assistant Professor of Management and Director of Artificial Intelligence Integration, Elon University.

Corresponding author: Mustafa Akben, 2075 Campus Box, Elon, NC 27244, United States. Email: makben@elon.edu

Leslie Coyne, Signalead LLC. Email: leslie.coyne@signalead.ai

## Abstract

AI is changing what leaders must judge, explain, learn, and coordinate, yet existing measures do not capture these behaviors at the level needed to study leadership in AI-enabled work. We develop the AI Leadership Battery, which organizes 36 behaviorally specific subdimensions into 11 theory-specified content families. Following established scale-development procedures, the research used deductive item generation; content validation of definitional correspondence and definitional distinctiveness; exploratory factor analysis and item reduction; confirmatory factor analysis in independent samples; and tests of internal consistency reliability, convergent validity, discriminant validity, and criterion-related and incremental validity. Across the development and validation studies, the analyses provided evidence for the Battery's content, multidimensional structure, reliability, and distinction from selected orbiting constructs. The Battery also contributed additional information beyond orbiting constructs across organizational growth, decision speed, customer/stakeholder response capability, AI-enabled team performance, AI-enabled work experience, AI security and risk management, and AI adoption and integration. The resulting measure provides researchers with a behavioral framework for examining how leaders in AI-enabled work regulate judgment, learning, adaptation, transparency, and accountability.

*Keywords:* artificial intelligence; leadership; scale development; psychometrics; human-AI collaboration; measurement

**Measuring AI Leadership: Development and Validation of a Behavioral Battery for AI-Native Organizations**

Artificial intelligence can change both the content of managerial work and the mechanisms through which leadership is enacted. E-leadership research indicates that advanced information technology does more than transmit influence: technology and leadership processes reciprocally shape one another across people, time, distance, and organizational structure (Avolio et al., 2014). We propose that AI may intensify this reciprocity by redistributing information, expertise, and decision input. In AI-enabled work, leaders decide when to rely on AI, when to challenge it, how to redesign roles, how to create safe learning conditions, how to explain AI-supported decisions, and where human responsibility remains.

The automation-augmentation paradox provides a framework for this problem. Automation and augmentation are interdependent choices that unfold across tasks and over time (Raisch & Krakowski, 2021). Automating routine output may increase the need for human interpretation, exception handling, and accountability. Leading AI-enabled work may therefore involve ongoing adjustment of leadership behavior rather than a generally favorable attitude toward technology.

Recent reviews describe AI-related leadership as broad and fragmented, spanning communication, adaptability, collaboration, ethics, human-AI interaction, implementation, algorithmic control, and organizational transformation (Aziz et al., 2025; Bankins et al., 2024; Bock & von der Oelsnitz, 2025). Recent studies illustrate narrower mechanisms. Leader-AI collaboration has been associated with team performance through reflexivity and team role breadth self-efficacy; leader AI crafting has been associated with employee crafting, engagement, and robot-directed helping; and algorithmic authority received lower fairness, trustworthiness, and legitimacy judgments than human authority in a moral-leadership experiment (Li et al., 2024; Liu et al., 2026; McGuire & De Cremer, 2023). Together, these studies reveal a common leadership problem: AI changes how influence is exercised through judgment, collaboration, learning, implementation, and accountability.

### Measurement Gap and Neighboring Domains

Existing measures assess related constructs. Human-centric digital leadership scales assess broad digital-transformation readiness, including ethical AI use, skills acquisition, and participative style (Abbu et al., 2025). Algorithmic leadership and algorithmic-management measures treat the algorithm or control system as the source of direction, monitoring, scheduling, evaluation, or relational functions (Chang et al., 2026; Parent-Rocheleau et al., 2024). AI-attitude and literacy measures assess beliefs, anxiety, utility, knowledge, and technical capability (Liu et al., 2025; Park et al., 2024; Wang et al., 2023). Organizational AI-responsibility and firm AI-capability measures address system-level governance or organizational resources (Jaturat et al., 2026; Mikalef & Gupta, 2021). The present study focuses on behaviorally specific actions through which leaders guide people, judgment, learning, and implementation in AI-enabled work. These actions extend established research on learning orientation, adaptation, boundary spanning, transparency, coaching, team learning, psychological safety, meaningful work, ethical leadership, and implementation leadership. AI changes the object, tradeoff, timing, and accountability structure of these behaviors: leaders must decide when to rely on AI, when to challenge it, how to organize work around it, and how to remain responsible for its consequences. The Battery brings these actions into a measurement framework for studying leadership in AI-enabled work.

**Theoretical Derivation of the Domain**

AI leadership refers to behaviorally specific actions through which leaders guide people, judgment, learning, and implementation in AI-enabled work. The framework differentiates these actions around three recurring conditions. First, probabilistic output, reliance on AI for cognitive tasks, and rapid knowledge change make expertise provisional and require leaders to continually evaluate what they know. Second, AI reconfigures tasks, roles, voice, learning, and work meaning, requiring leaders to coordinate how people adapt and develop. Third, AI redistributes decision input while leaving leaders responsible for implementation and consequences. These conditions provide the theoretical foundation for the Battery's 11 content families and 36 behavioral subdimensions. Table 1 summarizes the proposed links among the conditions, leadership functions, and content families.

Table 1

*From AI-enabled work conditions to theory-specified content families*

| **AI-enabled work condition** | **Leadership function** | **Content families** | **Representative literature** |
|---|---|---|---|
| Uncertainty in AI output and changing knowledge: probabilistic output, reliance on AI for cognitive tasks, and rapidly changing expertise | Practice updating, provisional action, and knowledge acquisition beyond familiar domains | Developmental Stretch; Adaptive Leadership; Outside-In Learning | Learning goal orientation; intellectual humility; adaptive leadership behavior; boundary spanning; knowledge brokering |
| Sociotechnical role change: altered tasks, capability demands, risks associated with speaking up, and work meaning | disclosure of AI involvement, capability development, learning routines, interpersonal safety, and shared meaning | Proactive Transparency; Role & Skill Evolution; Team Learning Experiences; Psychologically Safe Climate; Purposeful Communication | Transparency and informational justice; coaching; team learning and reflexivity; psychological safety; meaningful work |

| AI-enabled work condition | Leadership function | Content families | Representative literature |
|---|---|---|---|
| Human accountability demands as AI expands decision input | Reliance calibration, ethical boundary enforcement, and governance of implementation over time | Contextual Calibration; Ethical Guardrails & Accountability; Change Discipline | Trust in automation; actively open-minded thinking; ethical leadership; accountability; implementation and sustainment leadership |

The 36 subdimensions are deductively specified behavioral facets within the content families. They were defined before item writing, refined through development cycles, evaluated through content validation, and evaluated using factor analysis within each content family. This sequence preserves the connection between the theoretical domain, the behavioral definitions, the resulting items, and the structural evidence evaluated across the studies.

***Epistemic Judgment and Self-directed Adaptation***

Developmental Stretch refers to intentional capability development, updating practice from AI-related feedback, and receptiveness when AI challenges prior expertise. Its subdimensions are AI Developmental Stretch Orientation, AI Feedback-Driven Updating, and Expert Identity Flexibility. The family applies learning goal orientation, intellectual humility, and trust calibration to the ways leaders develop capabilities, revise their practices, and respond when AI challenges prior expertise (Jian et al., 2000; Krumrei-Mancuso & Rouse, 2016; VandeWalle, 1997).

Adaptive Leadership specifies provisional action and revision of AI-use decisions when evidence changes. Its subdimensions are Decisive Provisional Action Under AI Uncertainty, Evidence-Responsive Recalibration, and AI-Use Reconfiguration When Approaches Misfit. The subdimensions extend adaptive-leadership research to the decision cycle created by AI: acting under uncertainty, monitoring changing evidence, and reconfiguring AI use when an approach no longer fits the situation (Morrison & Phelps, 1999; Nöthel et al., 2023).

Outside-In Learning specifies scanning, brokering, and translating AI developments from beyond the team's usual domain. Its subdimensions are External AI Signal Scanning, External AI Perspective Brokering, and Cross-Domain AI Translation. This family contextualizes boundary spanning and knowledge brokering around rapidly evolving AI practice (Ancona & Caldwell, 1992; van Zoonen & Sivunen, 2023).

### *Social Coordination and Capability Development*

Proactive Transparency concerns disclosure of AI involvement before others must infer it, explanation of the reasons for AI use, and communication of limitations or harms. Its subdimensions are Proactive AI Role Disclosure, AI Use Rationale, and Candid AI Communication. In this family, AI involvement is the object of disclosure, including the visibility of system contributions to decisions (Colquitt, 2001; Schnackenberg et al., 2021).

Role & Skill Evolution specifies continuous development of the human-AI skill mix, redesign of work toward human value, and coaching of critical discernment. Its subdimensions are Continuous Human-AI Capability Development, Role Redesign Toward Human Value, and Coaching Critical AI Discernment. In this context, the family addresses both the allocation of work between people and systems and coaching for critical evaluation of AI output (Amundsen & Martinsen, 2014; Heslin et al., 2006; Wang et al., 2023).

Team Learning Experiences specifies recurring AI experimentation, capability review, transfer of learning into practice, public role modeling, and developmental handling of AI-related mistakes. Its subdimensions are Structured AI Experimentation and Learning Routines,

Systematic Applied Capability Review, Translating AI Learning into Ongoing Work Practices, Public Role Modeling, and Developmental Handling of AI-Related Mistakes (Edmondson, 1999; Schippers et al., 2007; van Dyck et al., 2005).

Psychologically Safe Climate specifies leader behaviors that support safety for AI experimentation, help seeking, dissent about AI output, and vulnerability about AI-driven role change. Its subdimensions are Safe AI Experimentation, AI Help-Seeking, Open AI Voice and Dissent, and Vulnerability About AI-Driven Change. The family applies psychological-safety research to the leader actions that allow team members to experiment with AI, request help, question its output, and discuss vulnerability associated with AI-driven role change (Edmondson, 1999).

Purposeful Communication specifies leader facilitation of collective interpretation of AI-related change and preservation of human purpose as roles evolve. Its subdimensions are Collective AI Sensemaking and Human Meaning Preservation. The family contextualizes meaningful-work and meaning-based leadership processes (Steger et al., 2012; van Knippenberg, 2020).

### *Accountable Implementation*

Contextual Calibration concerns situation-specific reliance decisions, questioning of AI assumptions, fit checking, and integration of local organizational context. Its subdimensions are Context-Sensitive Reliance Calibration, Critical Questioning of AI Assumptions, Deliberative Contextual Fit-Checking, and Integration of Local Organizational Context. These subdimensions make calibration observable through leaders’ continuing evaluation of evidence, assumptions, task requirements, organizational context, and the consequences of relying on AI (Haran et al., 2013; Jian et al., 2000).

Ethical Guardrails & Accountability specifies ethical boundary setting, enforcement despite cost, and personal answerability for AI-influenced outcomes. Its subdimensions are Ethical AI Boundary Setting, Costly Ethical Enforcement, and Personal Accountability for AI Outcomes. AI can alter decision-authority boundaries and may facilitate the deflection of responsibility to systems or vendors (Brown et al., 2005; Hall et al., 2003).

Change Discipline specifies pacing implementation to team capacity, governing continuation or change through explicit criteria, and coordinating sustainment. Its subdimensions are Absorptive Pacing and Responsiveness, Implementation Governance and Adaptive Evaluation, and Coordinated Sustainment. The family contextualizes implementation and sustainment leadership around evolving AI use (Aarons et al., 2014; Ehrhart et al., 2018).

## Contextualization and Construct Novelty

Psychological safety, for example, concerns voice about opaque output, role threat, and experimentation with probabilistic systems; coaching concerns human-AI discernment and role redesign; ethical leadership concerns continued responsibility when cognitive tasks are delegated to AI; and implementation leadership concerns governing systems whose performance and risks evolve through use. Together, these contextualized behaviors allow researchers to distinguish how leaders manage AI-related judgment, learning, role change, communication, and implementation. Their value lies in making specific leadership choices visible within a shared framework rather than treating AI leadership as a broad orientation toward technology.

## Measurement Framework and Terminology

In this manuscript, AI leadership is the broad behavioral domain under study. The 11 content families organize theoretically related leadership behaviors. Each family contains behavioral subdimensions, each measured by a three-item reflective scale. Form refers to one of three selected-item sets modeled within its corresponding survey sample. The labels 30-item, 51-item, and 27-item identify the modeled subsets, not the full administered survey lengths. Signal refers to a practitioner communication grouping that connects related content families. Behavioral refers to the action content and leader referent of the items, allowing the Battery to specify what leaders do across AI-enabled decisions, relationships, and work practices. Figure 1 displays the conceptual organization of the content families and their subdimensions.

**Figure 1**

*Conceptual Organization of the AI Leadership Battery*

**Inner Signal**
Learning and Adaptive Orientation

**Developmental Stretch**
AI Developmental Stretch Orientation
AI Feedback-Driven Updating
Expert Identity Flexibility

**Adaptive Leadership**
AI-Use Reconfiguration When Approaches Misfit
Decisive Provisional Action Under AI Uncertainty
Evidence-Responsive Recalibration

**Outside-In Learning**
Cross-Domain AI Translation
External AI Perspective Brokering
External AI Signal Scanning

**Outward Signal**
Interpersonal and Team Leadership Practices

**Proactive Transparency**
AI Use Rationale
Candid AI Communication
Proactive AI Role Disclosure

**Role & Skill Evolution**
Coaching Critical AI Discernment
Continuous Human-AI Capability Development
Role Redesign Toward Human Value

**Team Learning Experiences**
Developmental Handling of AI-Related Mistakes
Public Role Modeling
Structured AI Experimentation and Learning Routines
Systematic Applied Capability Review
Translating AI Learning into Ongoing Work Practices

**Psychologically Safe Climate**
AI Help-Seeking
Open AI Voice and Dissent
Safe AI Experimentation
Vulnerability About AI-Driven Change

**Purposeful Communication**
Collective AI Sensemaking
Human Meaning Preservation

**Grounded Signal**
Contextual Judgment and Implementation Governance

**Contextual Calibration**
Context-Sensitive Reliance Calibration
Critical Questioning of AI Assumptions
Deliberative Contextual Fit-Checking
Integration of Local Organizational Context

**Ethical Guardrails & Accountability**
Costly Ethical Enforcement
Ethical AI Boundary Setting
Personal Accountability for AI Outcomes

**Change Discipline**
Absorptive Pacing and Responsiveness
Coordinated Sustainment
Implementation Governance and Adaptive Evaluation

*Note.* The 11 content families comprise 36 subdimensions, each measured by three retained items. Inner, Outward, and Grounded Signals provide conceptual organization. The original validation groups comprised Adaptive Leadership, Outside-In Learning, and Change Discipline (27 retained items; n = 201); the five Outward families (51 retained items; n = 202); and Developmental Stretch, Contextual Calibration, and Ethical Guardrails & Accountability (30 retained items; n = 198). Developmental Stretch and Change Discipline occupy different groupings in the later conceptual framework. Statistical estimates refer to the original groups; item counts identify retained subsets of the administered surveys.

The research forms and practitioner Signals organize the framework for different purposes. The 30-item form includes Developmental Stretch, Contextual Calibration, and Ethical Guardrails & Accountability. The 51-item form includes the five Outward content families. The 27-item form includes Adaptive Leadership, Outside-In Learning, and Change Discipline. The practitioner framework recombines these content families into Inner, Outward, and Grounded Signals to communicate how leaders orient to AI, guide people through AI-enabled work, and maintain contextual and ethical grounding. Table 2 summarizes the samples, analyses, and validation objectives across the research program.

Table 2

*Research Samples, Analyses, and Validation Objectives*

| Stage | N | Analysis | Purpose of analysis |
|---|---|---|---|
| Content validity | 199 | Definitional correspondence and distinctiveness assessment | Evaluate how well candidate items represent their intended definitions |
| Exploratory factor analysis | 590 | EFA, item selection, and internal-consistency analysis | Examine subdimension structure and refine the item sets |
| Confirmatory factor analysis | 601 | CFA in independent validation samples | Evaluate the prespecified measurement structures |
| Convergent validity | 601 | Standardized loadings and average variance extracted | Evaluate convergence among indicators of each subdimension |
| Discriminant validity | 601 | HTMT, bootstrap intervals, and focal-side AVE comparisons | Evaluate separation from related comparison constructs, referred to here as orbiting constructs |
| Criterion-related and incremental validity | 601 | Hierarchical regression and repeated cross-validation | Examine concurrent outcome associations and added explanatory value beyond orbiting constructs |

*Note.* Development comprises 590 records from 589 unique leaders. Validation comprises 601 unique leaders with no overlap with development. The 199 panel records are analytic records and do not represent a verified unique-person count. The repeated N = 601 refers to the same three validation samples across analyses, not additional recruitment. Analyses were conducted separately by form; model-specific Ns appear in the detailed tables.

**Practitioner Framework**

For practitioner communication, Inner groups Developmental Stretch, Adaptive Leadership, and Outside-In Learning; Outward groups Proactive Transparency, Role & Skill Evolution, Team Learning Experiences, Psychologically Safe Climate, and Purposeful Communication; and Grounded groups Contextual Calibration, Ethical Guardrails & Accountability, and Change Discipline. This organization translates the Battery's research framework into a practical vocabulary. Inner describes how leaders develop and adapt their own orientation toward AI. Outward describes how leaders create the social and developmental conditions for AI-enabled work. Grounded describes how leaders calibrate AI use, establish boundaries, and maintain accountability during implementation.

This study makes two primary contributions. First, it defines AI leadership as a behavioral domain organized around the recurring demands of AI-enabled work, rather than as a single, generalized leadership attribute. Second, it provides initial measurement evidence for 36 behavioral subdimensions, each measured by three items, spanning that domain.

The present research program addresses four questions. First, do the candidate items align with their prespecified behavioral definitions? Second, within each content-family pool, do the proposed subdimensions demonstrate the intended internal structure and adequate internal consistency? Third, are the resulting subdimension scores empirically distinguishable from selected orbiting constructs included in the administered forms? Fourth, do the subdimensions contribute additional variance in outcomes measured in the same survey beyond that provided by the selected orbiting constructs?

These questions are examined sequentially across content-validation, development, and independent confirmation samples. The studies move from evaluating correspondence between items and behavioral definitions to examining exploratory structure, confirmatory structure, relations with established measures, and criterion-related and incremental validity. Each stage addresses a distinct part of the measurement argument while building evidence for the Battery's use in research on AI-enabled leadership.

The program used a deductive approach to scale development described in organizational measure-development guidance (Clark & Watson, 2019; Hinkin, 1998). In this manuscript, validity refers to interpretations and uses of scores rather than a permanent property of an instrument (American Educational Research Association et al., 2014). Definitions preceded item writing within development cycles; content review assessed definitional correspondence; exploratory analyses evaluated internal structure and supported item reduction; newly recruited samples were reserved for form-specific confirmatory models and selected external relations; and those validation samples were reused for descriptive same-wave criterion analyses. Following quantitative-reporting guidance, the manuscript connects each validity question to the sample, measure, and analysis used to examine it (Appelbaum et al., 2018). Together, the stages evaluate whether the Battery represents differentiated leadership behaviors and contributes information beyond established measures. Table 3 presents the behavioral subdimensions and item counts from candidate development through final retention.

Table 3

*Behavioral Subdimensions and Item Retention*

| Content family | Subdimension | Candidate items | Items evaluated in EFA | Final retained items |
|---|---|---|---|---|
| Developmental Stretch | AI Developmental Stretch Orientation | 5 | 5 | 3 |
| Developmental Stretch | AI Feedback-Driven Updating | 5 | 5 | 3 |
| Developmental Stretch | Expert Identity Flexibility | 5 | 5 | 3 |
| Adaptive Leadership | AI-Use Reconfiguration When Approaches Misfit | 5 | 5 | 3 |
| Adaptive Leadership | Decisive Provisional Action Under AI Uncertainty | 5 | 5 | 3 |
| Adaptive Leadership | Evidence-Responsive Recalibration | 5 | 5 | 3 |
| Outside-In Learning | Cross-Domain AI Translation | 5 | 5 | 3 |
| Outside-In Learning | External AI Perspective Brokering | 5 | 5 | 3 |
| Outside-In Learning | External AI Signal Scanning | 5 | 5 | 3 |
| Proactive Transparency | AI Use Rationale | 5 | 5 | 3 |
| Proactive Transparency | Candid AI Communication | 5 | 5 | 3 |
| Proactive Transparency | Proactive AI Role Disclosure | 5 | 5 | 3 |
| Role & Skill Evolution | Coaching Critical AI Discernment | 5 | 5 | 3 |
| Role & Skill Evolution | Continuous Human-AI Capability Development | 5 | 5 | 3 |
| Role & Skill Evolution | Role Redesign Toward Human Value | 5 | 5 | 3 |
| Team Learning Experiences | Developmental Handling of AI-Related Mistakes | 5 | 5 | 3 |
| Team Learning Experiences | Public Role Modeling | 5 | 5 | 3 |
| Team Learning Experiences | Structured AI Experimentation and Learning Routines | 5 | 5 | 3 |
| Team Learning Experiences | Systematic Applied Capability Review | 5 | 5 | 3 |
| Team Learning Experiences | Translating AI Learning into Ongoing Work Practices | 5 | 5 | 3 |
| Psychologically Safe Climate | AI Help-Seeking | 5 | 5 | 3 |
| Psychologically Safe Climate | Open AI Voice and Dissent | 5 | 5 | 3 |

| Content family | Subdimension | Candidate items | Items evaluated in EFA | Final retained items |
|---|---|---|---|---|
| Psychologically Safe Climate | Safe AI Experimentation | 5 | 5 | 3 |
| Psychologically Safe Climate | Vulnerability About AI-Driven Change | 5 | 5 | 3 |
| Purposeful Communication | Collective AI Sensemaking | 5 | 5 | 3 |
| Purposeful Communication | Human Meaning Preservation | 5 | 5 | 3 |
| Contextual Calibration | Context-Sensitive Reliance Calibration | 5 | 5 | 3 |
| Contextual Calibration | Critical Questioning of AI Assumptions | 5 | 5 | 3 |
| Contextual Calibration | Deliberative Contextual Fit-Checking | 5 | 5 | 3 |
| Contextual Calibration | Integration of Local Organizational Context | 5 | 5 | 3 |
| Ethical Guardrails & Accountability | Costly Ethical Enforcement | 5 | 5 | 3 |
| Ethical Guardrails & Accountability | Ethical AI Boundary Setting | 5 | 5 | 3 |
| Ethical Guardrails & Accountability | Personal Accountability for AI Outcomes | 5 | 5 | 3 |
| Change Discipline | Absorptive Pacing and Responsiveness | 5 | 5 | 3 |
| Change Discipline | Coordinated Sustainment | 5 | 5 | 3 |
| Change Discipline | Implementation Governance and Adaptive Evaluation | 5 | 5 | 3 |
| Total: 11 content families | 36 subdimensions | 180 | 180 | 108 |

*Note.* Candidate-item counts refer to the item pools following revision and content validation. All 180 candidate items were evaluated in exploratory factor analysis. The final Battery contains 108 items, with three items per subdimension.

## Stage 1: Content Definition and Content-Validity Assessment

### Item Generation and Refinement

The research team specified 11 content families and 36 subdimensions before writing items. Five candidate items were generated for each subdimension, producing 180 candidate items. Item generation emphasized direct descriptions of leader action and comprehensive coverage of each behavioral definition. The preprint presents the measurement framework, development procedures, decision rules, and summary results, while exact candidate and selected-item wording remains in the protected study materials.

Content review informed a second development cycle for AI Use Rationale and Structured AI Experimentation and Learning Routines. For AI Use Rationale, the revised items strengthened correspondence with the intended definition relative to the definitions of orbiting constructs. For Structured AI Experimentation and Learning Routines, the revisions improved coverage of the intended subdimension's definition. Newly recruited panels evaluated both revised item pools under the same protocol and comparison sets. The results below reflect the item pools retained through these development cycles.

**Panel Design and Participants**

Content validity was evaluated through 12 online content-validation surveys designed to assess definitional correspondence and definitional distinctiveness. Team Learning Experiences was divided across two surveys because it contained five subdimensions; each of the other content families used one survey. Each survey presented the intended subdimension definitions alongside two to five definitions of established orbiting constructs. Raters evaluated every candidate item against every definition in the applicable survey, producing a fully crossed within-rater design that assessed both correspondence with the intended definition and separation from the definitions of orbiting constructs.

Paid panelists were recruited through Prolific. Eligibility required a doctoral degree, residence in the United States, United Kingdom, or Canada, English as a first language, and professional or industry experience. Across the 12 content-validation panels, the analyses included 199 panel-level records. This total counts participation within panels rather than verified unique individuals. Prolific compensation was provided. Data were collected in June and July 2026.

**Procedure and Analysis**

The content-validation task used an item-rating approach informed by Hinkin and Tracey (1999) and the distinction between definitional correspondence and definitional distinctiveness described by Colquitt et al. (2019). The preprint summarizes the task structure and rating process, while the exact instructions, practice materials, and stimuli remain in the protected study materials. Raters used a seven-point rating scale to evaluate how well each candidate item matched each definition.

Three study-specific criteria were used to evaluate candidate items. First, the intended definition had to receive the highest mean rating among the definitions shown. Second, the intended-definition mean had to be at least 5.00. Third, the intended definition had to appear alone in the highest repeated-measures Duncan grouping relative to the published neighboring comparison definitions (Duncan, 1955), following the study's adaptation of content-adequacy procedures (Anderson & Gerbing, 1991; Hinkin & Tracey, 1999). Comparisons with other subdimensions within the same content family served as internal diagnostics. Candidate items were classified according to whether they met all three study-specific criteria. All candidate items advanced to exploratory factor analysis. This approach combined evidence of correspondence, distinctiveness, and later structural performance rather than allowing any single content statistic to determine retention.

**Results**

Of the 180 candidate items, 174 (96.7%) met all three study-specific criteria; six did not meet all three. Each of the 36 subdimensions had at least three candidate items that met all three criteria. All 180 candidate items advanced to exploratory factor analysis, with final item selection informed by both content-validity evidence and factor-analytic results. Table 3 summarizes the candidate counts and subsequent item retention. Table 4 provides a cross-stage summary of the retained items' exploratory and confirmatory loadings, described in Stages 2 and 3.

Table 4

*Exploratory and Confirmatory Factor Loadings*

| Content family / subdimension | Retained item numbers | EFA 1 | EFA 2 | EFA 3 | CFA 1 | CFA 2 | CFA 3 |
|---|---|---|---|---|---|---|---|
| **Developmental Stretch** | | | | | | | |
| AI Developmental Stretch Orientation | 1, 2, 3 | .664 | .760 | .822 | .74 | .92 | .92 |
| AI Feedback-Driven Updating | 1, 2, 5 | .837 | .802 | .900 | .93 | .91 | .88 |
| Expert Identity Flexibility | 2, 4, 5 | .671 | .736 | .691 | .68 | .83 | .86 |
| **Adaptive Leadership** | | | | | | | |
| AI-Use Reconfiguration When Approaches Misfit | 1, 2, 3 | .691 | .861 | .852 | .79 | .85 | .88 |
| Decisive Provisional Action Under AI Uncertainty | 1, 2, 3 | .901 | .859 | .779 | .85 | .90 | .91 |
| Evidence-Responsive Recalibration | 1, 2, 3 | .535 | .921 | .832 | .80 | .85 | .89 |
| **Outside-In Learning** | | | | | | | |
| Cross-Domain AI Translation | 1, 2, 3 | .483 | .928 | .717 | .76 | .87 | .92 |
| External AI Perspective Brokering | 1, 2, 4 | .732 | .893 | .725 | .89 | .91 | .82 |
| External AI Signal Scanning | 1, 2, 3 | .898 | .959 | .757 | .84 | .91 | .93 |
| **Proactive Transparency** | | | | | | | |
| AI Use Rationale | 1, 2, 3 | .668 | .690 | .872 | .77 | .75 | .75 |
| Candid AI Communication | 1, 2, 3 | .813 | .840 | .689 | .85 | .72 | .78 |

| Content family / subdimension | Retained item numbers | EFA 1 | EFA 2 | EFA 3 | CFA 1 | CFA 2 | CFA 3 |
|---|---|---|---|---|---|---|---|
| Proactive AI Role Disclosure | 1, 2, 3 | .564 | .907 | .776 | .78 | .86 | .88 |
| **Role & Skill Evolution** | | | | | | | |
| Coaching Critical AI Discernment | 1, 2, 3 | .776 | .940 | .818 | .81 | .95 | .82 |
| Continuous Human-AI Capability Development | 1, 2, 4 | .962 | .806 | .768 | .81 | .88 | .80 |
| Role Redesign Toward Human Value | 2, 3, 5 | .693 | .707 | .895 | .79 | .87 | .59 |
| **Team Learning Experiences** | | | | | | | |
| Developmental Handling of AI-Related Mistakes | 1, 2, 3 | .906 | .893 | .660 | .82 | .84 | .83 |
| Public Role Modeling | 1, 2, 3 | .720 | .896 | .507 | .84 | .87 | .69 |
| Structured AI Experimentation and Learning Routines | 1, 2, 3 | .802 | 1.000 | .638 | .88 | .91 | .81 |
| Systematic Applied Capability Review | 1, 2, 3 | .618 | .650 | .645 | .89 | .86 | .85 |
| Translating AI Learning into Ongoing Work Practices | 1, 2, 3 | .867 | .850 | .740 | .87 | .81 | .86 |
| **Psychologically Safe Climate** | | | | | | | |
| AI Help-Seeking | 1, 2, 3 | .758 | .856 | .835 | .75 | .88 | .91 |
| Open AI Voice and Dissent | 1, 2, 3 | .748 | .818 | .828 | .65 | .78 | .75 |
| Safe AI Experimentation | 1, 2, 4 | .873 | .958 | .698 | .92 | .92 | .67 |
| Vulnerability About AI-Driven Change | 1, 2, 3 | .869 | .789 | .771 | .81 | .90 | .74 |
| **Purposeful Communication** | | | | | | | |
| Collective AI Sensemaking | 1, 2, 4 | .696 | .885 | .835 | .74 | .85 | .73 |
| Human Meaning Preservation | 1, 2, 3 | .810 | .949 | .499 | .80 | .80 | .76 |
| **Contextual Calibration** | | | | | | | |
| Context-Sensitive Reliance Calibration | 1, 2, 3 | .687 | .475 | .595 | .75 | .86 | .77 |
| Critical Questioning of AI Assumptions | 1, 3, 4 | .512 | .983 | .498 | .83 | .85 | .78 |
| Deliberative Contextual Fit-Checking | 1, 2, 3 | .916 | .834 | .772 | .76 | .82 | .77 |

| Content family / subdimension | Retained item numbers | EFA 1 | EFA 2 | EFA 3 | CFA 1 | CFA 2 | CFA 3 |
|---|---|---|---|---|---|---|---|
| Integration of Local Organizational Context | 1, 2, 4 | .841 | .744 | .874 | .73 | .76 | .75 |
| **Ethical Guardrails & Accountability** | | | | | | | |
| Costly Ethical Enforcement | 1, 2, 3 | .914 | .997 | .814 | .87 | .96 | .90 |
| Ethical AI Boundary Setting | 1, 2, 3 | .940 | .942 | .675 | .92 | .93 | .85 |
| Personal Accountability for AI Outcomes | 1, 2, 3 | .900 | .899 | .922 | .84 | .93 | .87 |
| **Change Discipline** | | | | | | | |
| Absorptive Pacing and Responsiveness | 1, 2, 4 | .838 | .893 | .492 | .73 | .83 | .63 |
| Coordinated Sustainment | 1, 2, 3 | .864 | .928 | .943 | .84 | .93 | .91 |
| Implementation Governance and Adaptive Evaluation | 1, 2, 3 | .619 | .950 | .757 | .91 | .89 | .73 |

*Note.* EFA = exploratory factor analysis; CFA = confirmatory factor analysis. Within each row, columns 1–3 correspond to the retained item numbers in the order listed. EFA loadings are from analyses of the retained items in the development samples; CFA loadings are from the independent validation samples. Full item wording remains in the protected study materials. Supplementary item-loading details provide cross-loadings, standard errors, and confidence intervals.

Rater judgments formed the primary content-validity evidence. By evaluating each candidate item against its intended definition and the definitions of orbiting constructs, raters provided evidence of both definitional correspondence and definitional distinctiveness. These judgments established the content basis for the subsequent structural analyses in leader samples.

## Stage 2: Exploratory Factor Analysis, Item Selection, and Internal Consistency

### Participants and Screening

Working leaders were recruited through Prolific in three separately recruited development samples. Eligibility required age 18 or older, responsibility for at least one direct report or working team, familiarity with the tools and technologies used by the team, and comfort completing the survey in English. The development forms contained 50, 85, or 45 candidate items. Initial sample sizes were 219, 231, and 201, respectively.

The prespecified screening protocol included completion status, completion times below 180 seconds, failure of at least three embedded attention checks, 30 or more consecutive identical responses to candidate items, and within-person response variability below the second percentile for the applicable form. Integrity checks assessed identifiers, duplicate records, and out-of-range responses. Sixty-one records were excluded, yielding analytic samples of 194, 203, and 193 (590 records from 589 unique leaders). One retained participant completed both the 85- and 45-item forms; analyses remained separate by form. Mean age ranged from 40.98 to 43.98 years, median leadership experience ranged from 7 to 9 years, and the median number of direct reports ranged from 5 to 8.

## Measures and Analytic Strategy

All candidate items used a seven-point agreement response format and were scored in the same direction. Candidate items were administered in three development surveys containing 50, 85, and 45 items. Exploratory factor analyses were conducted separately for the item pool within each of the 11 content families, each comprising two to five proposed subdimensions. The exploratory stage evaluated the proposed factor structure and informed the selection of three items to represent each subdimension.

Suitability for factor analysis was evaluated using Kaiser–Meyer–Olkin (KMO) measures of sampling adequacy for each content-family item pool, item-level measures of sampling adequacy, and Bartlett's tests of sphericity. Factor retention was guided by common-factor parallel analysis using 1,000 random datasets and inspection of scree plots (Horn, 1965). Exploratory factor analyses used principal axis factoring with oblimin rotation, allowing factors to correlate (Fabrigar et al., 1999). Item selection considered factor loadings, cross-loadings, communalities, Stage 1 content-validity evidence, and coverage of each subdimension's behavioral definition.

Internal consistency was estimated separately for each three-item subdimension using coefficient alpha and standardized omega total. Mean inter-item correlations and corrected item-total correlations were also reported (Dunn et al., 2014; McDonald, 1999). No responses were imputed.

**Results**

Factorability statistics supported analysis of all 11 content-family item pools. Overall KMO values ranged from .884 to .958; the minimum item-level measure of sampling adequacy was .801; and all Bartlett's tests of sphericity were significant at $p < .001$. Parallel analysis and inspection of scree plots both indicated the prespecified subdimension count for each construct.

In the initial models, 177 of 180 candidate items loaded most strongly on the intended factor. Twelve unique candidate items met at least one of the study's loading, cross-loading, separation, or communality criteria for closer review. Selection combined these diagnostics with content-validity evidence and behavioral coverage to retain three items per subdimension, producing the 108-item Battery. In the EFAs of the retained items, all 108 retained items loaded most strongly on their intended factor. Maximum factor correlations within content families ranged from .546 to .788. The resulting item sets carried all 36 proposed behavioral subdimensions forward to independent confirmatory testing. Table 4 reports the retained-item loadings alongside the independent confirmatory estimates.

All 36 three-item subdimensions met the prespecified internal-consistency review criteria. Alpha ranged from .752 to .941, omega total from .754 to .943, and corrected item-total correlations from .565 to .924. These results showed consistent relationships among the items within each subdimension. Together, the loading patterns and reliability estimates provided the empirical basis for evaluating the selected measures in newly recruited validation samples. For the independent validation samples examined in Stage 3, Table 5 reports subdimension descriptive statistics, reliability estimates, and convergent-validity evidence.

Table 5

*Subdimension Descriptive Statistics, Reliability, and Convergent Validity*

| Content family / subdimension | N | M | SD | α | ω (score) | Model ω | AVE |
|---|---|---|---|---|---|---|---|
| **Developmental Stretch** | | | | | | | |
| AI Developmental Stretch Orientation | 198 | 5.12 | 1.35 | .89 | .90 | .90 | .75 |
| AI Feedback-Driven Updating | 198 | 5.15 | 1.39 | .93 | .93 | .93 | .82 |
| Expert Identity Flexibility | 198 | 5.53 | 1.01 | .82 | .84 | .84 | .63 |
| **Adaptive Leadership** | | | | | | | |
| AI-Use Reconfiguration When Approaches Misfit | 201 | 5.37 | 1.30 | .87 | .87 | .88 | .70 |
| Decisive Provisional Action Under AI Uncertainty | 201 | 4.47 | 1.48 | .92 | .92 | .92 | .79 |
| Evidence-Responsive Recalibration | 201 | 5.39 | 1.22 | .88 | .89 | .89 | .72 |
| **Outside-In Learning** | | | | | | | |
| Cross-Domain AI Translation | 201 | 5.31 | 1.18 | .88 | .89 | .89 | .73 |
| External AI Perspective Brokering | 201 | 5.19 | 1.27 | .91 | .91 | .91 | .77 |
| External AI Signal Scanning | 201 | 5.31 | 1.33 | .92 | .92 | .92 | .80 |
| **Proactive Transparency** | | | | | | | |
| AI Use Rationale | 202 | 5.38 | 1.11 | .80 | .80 | .80 | .57 |
| Candid AI Communication | 202 | 5.27 | 1.19 | .82 | .83 | .82 | .61 |
| Proactive AI Role Disclosure | 202 | 5.12 | 1.36 | .87 | .88 | .88 | .71 |
| **Role & Skill Evolution** | | | | | | | |
| Coaching Critical AI Discernment | 202 | 5.86 | 1.17 | .89 | .89 | .90 | .74 |

| Content family / subdimension | N | M | SD | α | ω (score) | Model ω | AVE |
|---|---|---|---|---|---|---|---|
| Continuous Human-AI Capability Development | 202 | 5.07 | 1.27 | .87 | .87 | .87 | .69 |
| Role Redesign Toward Human Value | 202 | 5.35 | 1.08 | .79 | .81 | .80 | .58 |
| **Team Learning Experiences** | | | | | | | |
| Developmental Handling of AI-Related Mistakes | 202 | 5.05 | 1.29 | .87 | .87 | .87 | .69 |
| Public Role Modeling | 202 | 5.17 | 1.27 | .84 | .85 | .85 | .65 |
| Structured AI Experimentation and Learning Routines | 202 | 4.20 | 1.54 | .90 | .90 | .90 | .75 |
| Systematic Applied Capability Review | 202 | 4.67 | 1.52 | .90 | .90 | .90 | .75 |
| Translating AI Learning into Ongoing Work Practices | 202 | 5.03 | 1.34 | .89 | .89 | .89 | .72 |
| **Psychologically Safe Climate** | | | | | | | |
| AI Help-Seeking | 202 | 6.04 | 0.87 | .88 | .89 | .89 | .72 |
| Open AI Voice and Dissent | 202 | 5.75 | 0.90 | .76 | .77 | .77 | .53 |
| Safe AI Experimentation | 202 | 5.25 | 1.27 | .87 | .88 | .88 | .72 |
| Vulnerability About AI-Driven Change | 202 | 6.01 | 0.89 | .85 | .86 | .86 | .67 |
| **Purposeful Communication** | | | | | | | |
| Collective AI Sensemaking | 202 | 5.37 | 1.06 | .81 | .83 | .82 | .60 |
| Human Meaning Preservation | 202 | 5.29 | 1.15 | .83 | .83 | .83 | .62 |
| **Contextual Calibration** | | | | | | | |
| Context-Sensitive Reliance Calibration | 198 | 5.76 | 1.03 | .83 | .83 | .84 | .63 |
| Critical Questioning of AI Assumptions | 198 | 5.58 | 1.03 | .86 | .86 | .86 | .67 |
| Deliberative Contextual Fit-Checking | 198 | 5.87 | 0.87 | .83 | .83 | .83 | .62 |
| Integration of Local Organizational Context | 198 | 5.49 | 0.96 | .79 | .79 | .79 | .56 |
| **Ethical Guardrails & Accountability** | | | | | | | |
| Costly Ethical Enforcement | 198 | 5.42 | 1.44 | .94 | .94 | .94 | .83 |
| Ethical AI Boundary Setting | 198 | 5.18 | 1.49 | .92 | .93 | .93 | .81 |
| Personal Accountability for AI Outcomes | 198 | 5.52 | 1.25 | .91 | .91 | .91 | .77 |
| **Change Discipline** | | | | | | | |
| Absorptive Pacing and Responsiveness | 201 | 5.33 | 1.04 | .77 | .78 | .78 | .54 |
| Coordinated Sustainment | 201 | 4.82 | 1.49 | .92 | .92 | .92 | .79 |

| Content family / subdimension | N | M | SD | α | ω (score) | Model ω | AVE |
| --- | --- | --- | --- | --- | --- | --- | --- |
| Implementation Governance and Adaptive Evaluation | 201 | 4.96 | 1.33 | .87 | .88 | .88 | .72 |

*Note.* Statistics are from the independent validation samples. M = mean; SD = standard deviation; α = Cronbach's alpha; ω (score) = score-based omega; Model ω = omega reported by the confirmatory factor model; AVE = average variance extracted.

## Stage 3: Confirmatory Factor Analysis, Reliability, and Validity Evidence

### Participants and Screening

Three new samples of working leaders were recruited through Prolific using the same eligibility criteria. Each participant completed one validation survey. The validation surveys contained 50, 85, or 45 candidate items; the confirmatory factor analyses used the prespecified subsets of 30, 51, or 27 retained items, respectively. Initial sample sizes were 220, 225, and 222. Applying the prespecified screening criteria excluded 66 respondents and retained 198, 202, and 201, for a total validation sample of 601. Mean age ranged from 40.42 to 43.38 years, median leadership experience from 6 to 9 years, and median direct reports from 5 to 7.

### Measures

The three confirmatory factor models represented 10, 17, and 9 AI Leadership subdimensions, respectively, with three items per subdimension. All focal items were forward-keyed and used the seven-point agreement format. Eleven orbiting constructs were assessed using three-item, literature-based measures; each was administered only on the form to which it had been assigned on theoretical grounds. The 30-item form included Ethical Leadership (Brown et al., 2005), Learning Goal Orientation (VandeWalle, 1997), and Trust in Automation (Jian et al., 2000). The 51-item form included Informational Justice (Colquitt, 2001), Managerial Coaching (Heslin et al., 2006), Meaningful Work (Steger et al., 2012), Team Learning Behavior, and Team Psychological Safety (Edmondson, 1999). The 27-item form included Adaptive

Leadership Behavior (Nöthel et al., 2023), Boundary Spanning (Ancona & Caldwell, 1992), and Implementation Leadership (Aarons et al., 2014).

The source references for the orbiting constructs are identified above, and the administered item wording is documented in the study codebooks. All orbiting-construct items used the same seven-point agreement response format. The analyses examined relationships between these three-item measures and the AI Leadership subdimensions assessed on the same form.

Three color-preference items formed the marker variable used in a sensitivity analysis for common method variance. Subdimension and orbiting-construct scores were unweighted item means. No responses were imputed.

**Confirmatory Models**

One prespecified correlated-factor CFA model was estimated per form using robust maximum likelihood. Each item loaded on its hypothesized factor, cross-loadings were fixed to zero, factor variances were fixed to one for model identification, and factors within each form were allowed to correlate. The models contained 10 factors and 30 items, 17 factors and 51 items, and 9 factors and 27 items. The prespecified item assignments and model structure were maintained throughout the confirmatory analyses, providing an independent evaluation of the measures selected in Stage 2.

Model fit was evaluated using the robust comparative fit index (CFI), robust Tucker–Lewis index (TLI), robust root-mean-square error of approximation (RMSEA), and standardized root-mean-square residual (SRMR). These complementary indices were interpreted jointly (Marsh et al., 2004). Standardized loadings, composite reliability, and average variance extracted provided additional evidence about the measurement properties of each subdimension.

## Discriminant Validity Relative to Orbiting Constructs

Every AI Leadership subdimension was evaluated against every orbiting construct administered on the same form, yielding 142 evaluated relationships across the three samples. Three prespecified study criteria guided evaluation: a heterotrait–monotrait ratio of correlations (HTMT) below .85, an upper bound of the 95% bootstrap confidence interval for HTMT below .90, and a study-specific adaptation of the Fornell–Larcker criterion requiring the focal AI Leadership subdimension's average variance extracted (AVE) to exceed its squared observed correlation with the orbiting construct (Fornell & Larcker, 1981). Bootstrap intervals used 1,000 respondent-level resamples. Together, these procedures examined the magnitude and uncertainty of the relationships between the focal subdimensions and the administered orbiting constructs (Henseler et al., 2015; Rönkkö & Cho, 2022).

The marker-variable analysis examined whether the orbiting-construct results were sensitive to variance associated with the color-preference marker. We calculated partial correlations controlling for the marker variable and recalculated HTMT using partial item correlations, then compared the adjusted results with the unadjusted results against the prespecified criteria (Podsakoff et al., 2003).

## Confirmatory Results

All three models converged and were admissible. For the 30-item form, robust CFI = .95, robust TLI = .93, robust RMSEA = .058, 90% CI [.049, .067], and SRMR = .061. For the 51-item form, robust CFI = .89, robust TLI = .87, robust RMSEA = .062, 90% CI [.057, .067], and SRMR = .059. For the 27-item form, robust CFI = .98, robust TLI = .98, robust RMSEA = .033, 90% CI [.014, .046], and SRMR = .042. The 51-item model estimated 238 free parameters in a sample of 202. Global fit was stronger for the 30- and 27-item forms, whereas the 51-item form showed mixed evidence across fit indices. Table 6 summarizes fit for the three confirmatory models.

Across forms, standardized loadings ranged from .59 to .96. Composite reliability ranged from .77 to .94, and average variance extracted ranged from .53 to .83. These estimates met the study's prespecified subdimension-level benchmarks. The results provided evidence of reliable item sets and convergence among the indicators representing each behavioral subdimension in the independent validation samples. Table 4 reports the item-level CFA loadings, and Table 5 reports the subdimension-level descriptive, reliability, and convergent-validity statistics.

Table 6

*Confirmatory Factor Analysis Model Fit*

| **Measurement model** | **N** | **Parameters** | **Scaled $\chi^2$(df)** | **Robust CFI** | **Robust TLI** | **Robust RMSEA [90% CI]** | **SRMR** |
|---|---|---|---|---|---|---|---|
| 10-factor, 30-item model | 198 | 105 | 570.9 (360) | .95 | .93 | .058 [.049, .067] | .061 |
| 17-factor, 51-item model | 202 | 238 | 1832.9 (1088) | .89 | .87 | .062 [.057, .067] | .059 |
| 9-factor, 27-item model | 201 | 90 | 339.6 (288) | .98 | .98 | .033 [.014, .046] | .042 |

*Note.* The three correlated-factor models were estimated using robust maximum likelihood in independent validation samples. CFI = comparative fit index; TLI = Tucker–Lewis index; RMSEA = root-mean-square error of approximation; SRMR = standardized root-mean-square residual.

All 142 subdimension–comparison construct pairs met the three prespecified study criteria for discriminant validity. The maximum HTMT was .72, and the highest upper bound of the 95% bootstrap confidence intervals was .82. Every focal subdimension also met the adapted

AVE-versus-squared-correlation criterion. After adjustment for the marker variable, the maximum HTMT was .71 and all pairs continued to meet the prespecified HTMT criteria. These findings provided consistent evidence of discriminant validity relative to the orbiting constructs included within each research form. Table 7 identifies the orbiting construct with the highest HTMT for each subdimension.

Table 7

*Orbiting Construct with the Highest HTMT for Each AI Leadership Subdimension*

| Content family / subdimension | Closest orbiting construct | r [95% CI] | HTMT [95% CI] | AVE |
|---|---|---|---|---|
| **Developmental Stretch** | | | | |
| AI Developmental Stretch Orientation | Trust in Automation | .64 [.55, .71] | .72 [.59, .82] | .75 |
| AI Feedback-Driven Updating | Trust in Automation | .55 [.44, .64] | .61 [.45, .73] | .82 |
| Expert Identity Flexibility | Trust in Automation | .61 [.51, .69] | .71 [.56, .82] | .63 |
| **Adaptive Leadership** | | | | |
| AI-Use Reconfiguration When Approaches Misfit | Implementation Leadership | .28 [.15, .40] | .35 [.20, .52] | .70 |
| Decisive Provisional Action Under AI Uncertainty | Boundary Spanning | .18 [.04, .31] | .20 [.08, .36] | .79 |
| Evidence-Responsive Recalibration | Implementation Leadership | .29 [.16, .41] | .37 [.22, .54] | .72 |
| **Outside-In Learning** | | | | |
| Cross-Domain AI Translation | Boundary Spanning | .51 [.40, .61] | .58 [.42, .73] | .73 |
| External AI Perspective Brokering | Boundary Spanning | .57 [.47, .66] | .64 [.50, .75] | .77 |
| External AI Signal Scanning | Boundary Spanning | .54 [.43, .63] | .60 [.45, .72] | .80 |
| **Proactive Transparency** | | | | |
| AI Use Rationale | Managerial Coaching | .29 [.16, .41] | .37 [.22, .52] | .57 |
| Candid AI Communication | Informational Justice | .28 [.15, .40] | .41 [.26, .59] | .61 |
| Proactive AI Role Disclosure | Informational Justice | .29 [.16, .41] | .41 [.28, .55] | .71 |
| **Role & Skill Evolution** | | | | |
| Coaching Critical AI Discernment | Managerial Coaching | .35 [.22, .46] | .42 [.27, .56] | .74 |

| Content family / subdimension | Closest orbiting construct | r [95% CI] | HTMT [95% CI] | AVE |
|---|---|---|---|---|
| Continuous Human-AI Capability Development | Managerial Coaching | .30 [.17, .42] | .37 [.22, .52] | .69 |
| Role Redesign Toward Human Value | Managerial Coaching | .24 [.11, .37] | .32 [.17, .49] | .58 |
| **Team Learning Experiences** | | | | |
| Developmental Handling of AI-Related Mistakes | Managerial Coaching | .27 [.14, .39] | .33 [.18, .50] | .69 |
| Public Role Modeling | Team Psychological Safety | .33 [.20, .45] | .45 [.26, .62] | .65 |
| Structured AI Experimentation and Learning Routines | Informational Justice | .20 [.06, .33] | .28 [.17, .44] | .75 |
| Systematic Applied Capability Review | Team Learning Behavior | .27 [.13, .39] | .32 [.17, .47] | .75 |
| Translating AI Learning into Ongoing Work Practices | Managerial Coaching | .29 [.16, .41] | .36 [.20, .50] | .72 |
| **Psychologically Safe Climate** | | | | |
| AI Help-Seeking | Team Psychological Safety | .40 [.28, .51] | .54 [.34, .72] | .72 |
| Open AI Voice and Dissent | Managerial Coaching | .41 [.29, .52] | .53 [.39, .67] | .53 |
| Safe AI Experimentation | Team Psychological Safety | .39 [.27, .50] | .49 [.29, .68] | .72 |
| Vulnerability About AI-Driven Change | Team Psychological Safety | .47 [.35, .57] | .65 [.50, .82] | .67 |
| **Purposeful Communication** | | | | |
| Collective AI Sensemaking | Managerial Coaching | .43 [.30, .53] | .54 [.40, .67] | .60 |
| Human Meaning Preservation | Team Learning Behavior | .32 [.19, .43] | .39 [.23, .54] | .62 |
| **Contextual Calibration** | | | | |
| Context-Sensitive Reliance Calibration | Ethical Leadership | .46 [.34, .56] | .57 [.38, .72] | .63 |
| Critical Questioning of AI Assumptions | Ethical Leadership | .46 [.34, .56] | .55 [.40, .71] | .67 |
| Deliberative Contextual Fit-Checking | Ethical Leadership | .41 [.29, .52] | .51 [.32, .69] | .62 |
| Integration of Local Organizational Context | Ethical Leadership | .32 [.19, .44] | .41 [.24, .57] | .56 |
| **Ethical Guardrails & Accountability** | | | | |
| Costly Ethical Enforcement | Ethical Leadership | .49 [.38, .59] | .57 [.42, .71] | .83 |
| Ethical AI Boundary Setting | Ethical Leadership | .42 [.30, .53] | .48 [.37, .60] | .81 |
| Personal Accountability for AI Outcomes | Ethical Leadership | .42 [.30, .53] | .50 [.35, .62] | .77 |
| **Change Discipline** | | | | |
| Absorptive Pacing and Responsiveness | Implementation Leadership | .38 [.25, .49] | .51 [.30, .70] | .54 |

| Content family / subdimension | Closest orbiting construct | r [95% CI] | HTMT [95% CI] | AVE |
|---|---|---|---|---|
| Coordinated Sustainment | Implementation Leadership | .26 [.13, .39] | .32 [.18, .47] | .79 |
| Implementation Governance and Adaptive Evaluation | Boundary Spanning | .36 [.24, .48] | .41 [.27, .56] | .72 |

*Note.* The table reports the orbiting construct producing the highest HTMT for each subdimension. HTMT = heterotrait–monotrait ratio; AVE = average variance extracted. Complete results for all latent correlations and orbiting-construct relationships appear in the supplementary materials.

Within the AI Leadership CFA models, the largest latent factor correlations were .85, .81, and .79. These correlations involved subdimensions representing learning and adaptation behaviors and indicate substantial shared variance among some subdimensions. Future research can further examine the discriminant validity of these closely related subdimensions. The comparisons with orbiting constructs provide evidence of distinctiveness relative to the comparison measures administered in this study.

## Stage 4: Concurrent and Incremental Validity

### Design and Measures

This stage examined whether the AI Leadership Battery explained variation in work-related outcome scores beyond the orbiting constructs administered in Stage 3. Analyses used the three validation samples, with predictors and outcomes reported by the same leaders in the same survey wave. Models were estimated separately for each form. The 36 subdimension scores were calculated as unweighted means of their three items. Eleven content-family composites were calculated by averaging the relevant subdimension scores. Separate regression models added either the relevant content-family composites or the relevant subdimension scores to examine incremental validity at both levels.

The baseline model for each form included its study-specific three-item orbiting constructs. Seven multi-item outcomes were examined: organizational growth, decision speed, customer/stakeholder response capability, AI-enabled team performance, AI-enabled work experience, AI security and risk management, and AI adoption and integration. These outcomes connected the measurement program to organizational functioning, experiences of AI-enabled work, and AI implementation. They contained 4, 3, 4, 5, 4, 4, and 4 scored items, respectively, and were calculated as unweighted item means on seven-point response scales. The embedded attention-check item was excluded from organizational growth. Internal-consistency estimates (coefficient alpha and omega) ranged from .85 to .96 across surveys. The same outcome set was examined within each form, allowing the results to describe the range of concurrent relationships associated with its leadership behaviors.

**Analytic Strategy**

Hierarchical multiple regression using ordinary least squares (OLS) entered the orbiting-construct scores in the first block and added either the form's content-family composites or its subdimension scores in the second block. Change in $R^2$ ($\Delta R^2$) quantified the additional outcome variance explained by the AI Leadership scores, and F-tests of the change in $R^2$ compared the nested models. Separate composite and subdimension models examined whether broader summaries and more detailed behavioral scores added explanatory value beyond the same baseline. Supplementary commonality analysis partitioned explained variance into components unique to each predictor block and shared between them. Dominance analysis examined the relative importance of predictors in explaining outcome variance. The primary results emphasize blockwise increments, with the supplementary analyses clarifying how predictors contributed to the fitted models.

Repeated k-fold cross-validation evaluated predictive performance on observations excluded from model estimation. Within each form, five-fold cross-validation was repeated five times for the orbiting-construct baseline and the models adding content-family composites or subdimensions. Mean held-out $R^2$ summarized performance across the repeated folds. These estimates complemented the in-sample increments by evaluating how the fitted relationships performed on held-out responses from the same validation sample. The F-tests of the change in $R^2$ were reported without adjustment for multiple comparisons.

**Results**

In-sample $\Delta R^2$ ranged from .009 to .436 for content-family blocks and from .042 to .522 for subdimension blocks. Unadjusted F-tests of the change in $R^2$ indicated statistically significant increments in 19 of 21 content-family models and 18 of 21 subdimension models. The two nonsignificant content-family increments concerned decision speed in the 30- and 51-item forms. The largest increments occurred for AI adoption and integration in the 51-item sample: the content-family block added .436 and the subdimension block added .522 beyond a baseline $R^2$ of .071. For the content-family model, commonality analysis partitioned the combined $R^2$ of .507 into .436 unique to AI Leadership, .015 unique to the orbiting constructs, and .057 shared between the blocks. Thus, 85.9% of the variance explained by this fitted model was uniquely associated with the AI Leadership block. Table 8, Panel A, reports the increments, confidence intervals, and tests; Figure 2 displays the pattern across outcomes.

Adding AI Leadership scores increased mean held-out $R^2$ above the orbiting-construct baseline in 18 of 21 content-family models and 14 of 21 subdimension models. For AI adoption and integration in the 51-item form, mean held-out $R^2$ was .009 for the baseline, .430 with content-family composites, and .417 with subdimensions. Both specifications improved held-out performance for this outcome, with the content-family model producing the higher value. Across the models, improvement over the baseline occurred more frequently with content-family composites. These results describe both in-sample incremental validity and predictive performance on held-out responses for the two scoring approaches. Table 8, Panel B, and Figure 3 compare mean held-out performance across the baseline and both AI Leadership specifications.

Table 8

*Criterion-Related and Incremental Validity*

**Panel A. Incremental Variance Explained Beyond Orbiting Constructs**

*10-Factor, 30-Item Model*

| Outcome | N | Baseline $R^2$ | Content-family $\Delta R^2$ [95% CI] | *p* | Subdimension $\Delta R^2$ [95% CI] | *p* |
|---|---|---|---|---|---|---|
| Organizational Growth | 198 | .322 | .052 [.017, .128] | .002 | .088 [.053, .205] | .004 |
| Decision Speed | 198 | .052 | .009 [.001, .069] | .618 | .079 [.052, .213] | .089 |
| Customer/Stakeholder Response Capability | 198 | .380 | .026 [.007, .096] | .041 | .042 [.032, .162] | .212 |
| AI-Enabled Team Performance | 198 | .322 | .057 [.013, .143] | < .001 | .097 [.051, .214] | .001 |
| AI-Enabled Work Experience | 198 | .314 | .105 [.038, .196] | < .001 | .124 [.072, .257] | < .001 |
| AI Security and Risk Management | 198 | .333 | .089 [.024, .185] | < .001 | .112 [.059, .243] | < .001 |
| AI Adoption and Integration | 198 | .253 | .056 [.013, .151] | .002 | .103 [.060, .232] | .002 |

*17-Factor, 51-Item Model*

| Outcome | N | Baseline $R^2$ | Content-family $\Delta R^2$ [95% CI] | *p* | Subdimension $\Delta R^2$ [95% CI] | *p* |
|---|---|---|---|---|---|---|
| Organizational Growth | 202 | .183 | .103 [.051, .205] | < .001 | .128 [.109, .300] | .016 |
| Decision Speed | 202 | .065 | .042 [.014, .145] | .117 | .088 [.089, .272] | .361 |
| Customer/Stakeholder Response Capability | 202 | .218 | .076 [.032, .196] | .001 | .134 [.109, .310] | .006 |
| AI-Enabled Team Performance | 202 | .129 | .282 [.171, .416] | < .001 | .367 [.299, .525] | < .001 |
| AI-Enabled Work Experience | 202 | .081 | .277 [.170, .408] | < .001 | .337 [.260, .521] | < .001 |
| AI Security and Risk Management | 202 | .106 | .237 [.142, .369] | < .001 | .312 [.246, .477] | < .001 |
| AI Adoption and Integration | 202 | .071 | .436 [.313, .559] | < .001 | .522 [.441, .644] | < .001 |

*9-Factor, 27-Item Model*

| Outcome | N | Baseline $R^2$ | Content-family $\Delta R^2$ [95% CI] | *p* | Subdimension $\Delta R^2$ [95% CI] | *p* |
|---|---|---|---|---|---|---|
| Organizational Growth | 201 | .099 | .072 [.024, .184] | .001 | .090 [.061, .226] | .016 |
| Decision Speed | 201 | .089 | .071 [.017, .186] | .001 | .104 [.056, .248] | .006 |
| Customer/Stakeholder Response Capability | 201 | .238 | .065 [.025, .142] | < .001 | .147 [.083, .281] | < .001 |
| AI-Enabled Team Performance | 201 | .124 | .152 [.053, .303] | < .001 | .196 [.112, .371] | < .001 |
| AI-Enabled Work Experience | 201 | .082 | .150 [.060, .287] | < .001 | .209 [.130, .371] | < .001 |
| AI Security and Risk Management | 201 | .180 | .153 [.071, .274] | < .001 | .192 [.130, .353] | < .001 |
| AI Adoption and Integration | 201 | .089 | .129 [.049, .260] | < .001 | .179 [.105, .333] | < .001 |

*Note. Baseline models contain the orbiting constructs assigned to each validation sample. Content-family and subdimension entries report the additional variance explained by the respective AI Leadership blocks, with bootstrap percentile confidence intervals. The reported p values are from the corresponding unadjusted F-tests of the change in $R^2$.*

## Panel B. Repeated Cross-Validation

*10-Factor, 30-Item Model*

| Outcome | Baseline held-out $R^2$ | + Content families | + Subdimensions |
|---|---|---|---|
| Organizational Growth | .279 | .312 | .282 |
| Decision Speed | -.001 | -.025 | -.044 |
| Customer/Stakeholder Response Capability | .275 | .240 | .142 |
| AI-Enabled Team Performance | .236 | .264 | .241 |
| AI-Enabled Work Experience | .232 | .330 | .259 |
| AI Security and Risk Management | .273 | .340 | .297 |
| AI Adoption and Integration | .178 | .201 | .186 |

*17-Factor, 51-Item Model*

| Outcome | Baseline held-out $R^2$ | + Content families | + Subdimensions |
|---|---|---|---|
| Organizational Growth | .115 | .176 | -.003 |
| Decision Speed | -.016 | -.058 | -.200 |
| Customer/Stakeholder Response Capability | .150 | .155 | .080 |
| AI-Enabled Team Performance | .054 | .297 | .267 |
| AI-Enabled Work Experience | -.011 | .245 | .168 |
| AI Security and Risk Management | .043 | .245 | .209 |
| AI Adoption and Integration | .009 | .430 | .417 |

*9-Factor, 27-Item Model*

| Outcome | Baseline held-out $R^2$ | + Content families | + Subdimensions |
|---|---|---|---|
| Organizational Growth | .064 | .090 | .032 |
| Decision Speed | .046 | .086 | .042 |
| Customer/Stakeholder Response Capability | .207 | .247 | .245 |
| AI-Enabled Team Performance | .068 | .181 | .154 |
| AI-Enabled Work Experience | .037 | .147 | .147 |
| AI Security and Risk Management | .124 | .245 | .206 |
| AI Adoption and Integration | .041 | .133 | .100 |

*Note.* Entries are mean held-out $R^2$ values from five repetitions of five-fold cross-validation within each validation sample.

**Figure 2**

*Incremental Explained Variance Across Outcomes and Research Forms*

| | Inner Signal[1] Learning and Adaptive Orientation n = 201 | | Outward Signal Interpersonal and Team Leadership Practices n = 202 | | Grounded Signal[1] Contextual Judgment and Implementation Governance n = 198 | |
|---|---|---|---|---|---|---|
| | Content families | Subdimensions | Content families | Subdimensions | Content families | Subdimensions |
| Organizational growth | .072 | .090 | .103 | .128 | .052 | .088 |
| Decision speed | .071 | .104 | .042 | .088 | .009 | .079 |
| Customer/stakeholder response capability | .065 | .147 | .076 | .134 | .026 | .042 |
| AI-enabled team performance | .152 | .196 | .282 | .367 | .057 | .097 |
| AI-enabled work experience | .150 | .209 | .277 | .337 | .105 | .124 |
| AI security and risk management | .153 | .192 | .237 | .312 | .089 | .112 |
| AI adoption and integration | .129 | .179 | .436 | .522 | .056 | .103 |

.00 .10 .20 .30 .40 .50 .60
Incremental explained variance ($\Delta R^2$)

*Note.* Cells report in-sample $\Delta R^2$ after adding content-family composites or subdimension scores to the orbiting-construct baseline. Shading uses a common scale; values are rounded to three decimals. Confidence intervals and tests appear in Table 8, Panel A. [1] Panel labels follow the practitioner vocabulary, while estimates retain the original research groupings. The Inner-labeled panel includes Adaptive Leadership, Outside-In Learning, and Change Discipline (n = 201); the Grounded-labeled panel includes Developmental Stretch, Contextual Calibration, and Ethical Guardrails & Accountability (n = 198). The Outward panel includes its five content families (n = 202). In the practitioner framework, Developmental Stretch belongs to Inner and Change Discipline to Grounded. Values represent the original content-family or subdimension blocks, not aggregate Signal scores.

**Figure 3**

*Held-Out Model Performance Across Outcomes and Research Forms*

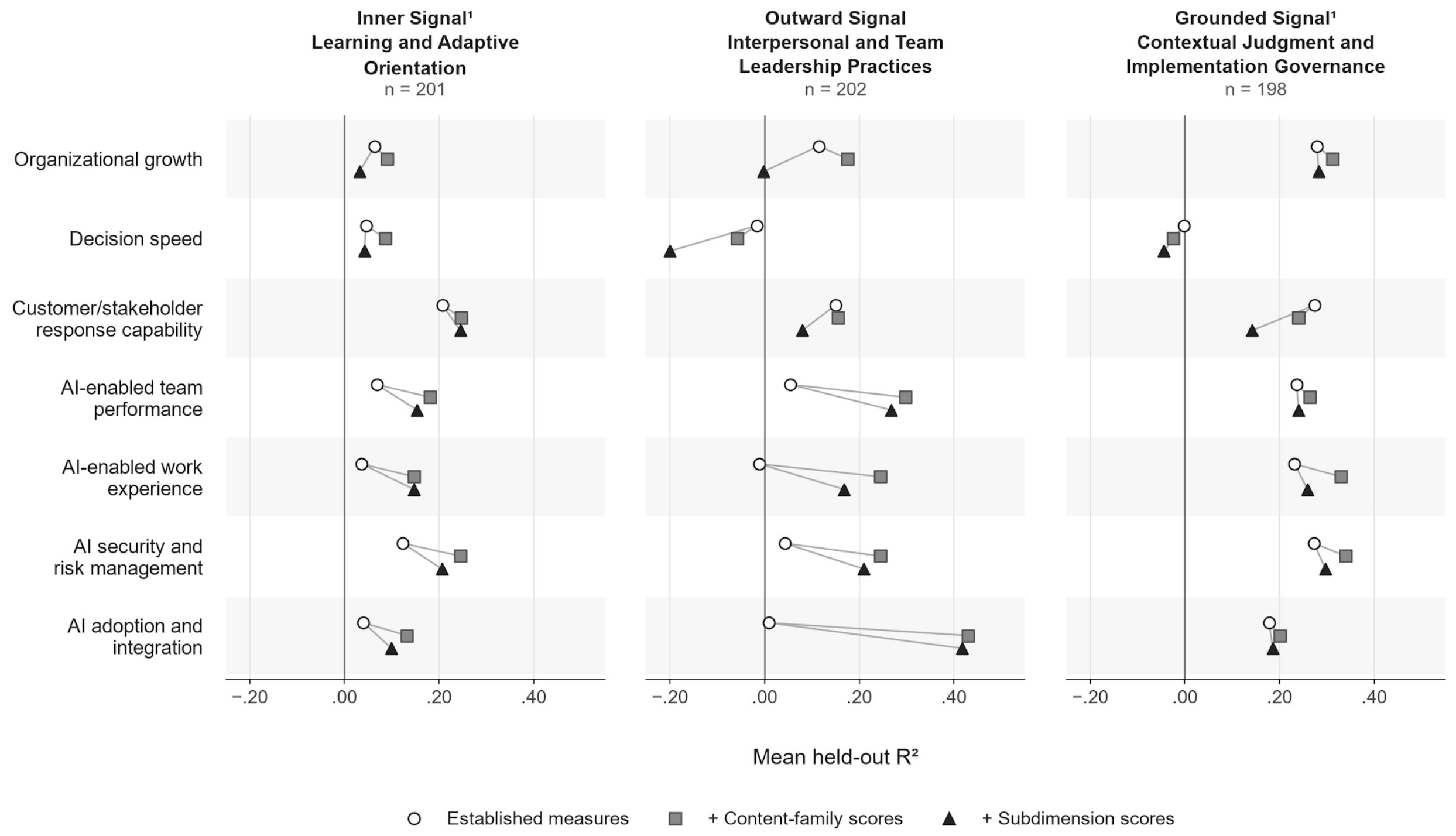

*Note.* Points report mean held-out $R^2$ from five repetitions of five-fold cross-validation (Table 8, Panel B). Open circles denote the orbiting-construct baseline; gray squares add content-family composites; black triangles add subdimension scores. Lines connect estimates for the same outcome; vertical offsets distinguish nearby markers. [1] Panel labels and research-group membership follow Figure 2. Estimates retain the original research groupings, including the placement of Change Discipline in the Inner-labeled sample and Developmental Stretch in the Grounded-labeled sample; they do not represent aggregate Signal scores.

Within the 51-item form, the largest increments concerned AI adoption and integration, AI-enabled team performance, and AI-enabled work experience. The 27-item form showed a more even distribution of increments across the outcomes, while the 30-item form yielded smaller increments. Each form used its own sample, behavioral content, and orbiting constructs. The results therefore identify where each set of leadership behaviors added explanatory value within its corresponding analysis. Taken together, the findings provide evidence of concurrent incremental validity for the measured organizational and AI-related outcomes.

# General Discussion

## Overview of Contributions

This research defines AI leadership through the specific actions leaders take to guide judgment, learning, relationships, and implementation in AI-enabled work. The AI Leadership Battery measures 36 behavioral subdimensions organized within 11 content families. The research program combines deductive construct definition and content validation with exploratory factor analysis and item selection, confirmatory factor analysis in independent samples, and analyses of concurrent and incremental validity. Together, these stages provide a theoretical organization of AI-related leadership behavior and initial measurement evidence for examining that behavior in organizational research.

The findings provide initial evidence of reliability and validity for the Battery's subdimension scores and their relationships with work outcomes. Across the validation samples, item loadings, composite reliability, and average variance extracted met the study's subdimension-level benchmarks. Relationships with the administered comparison measures provided evidence of discriminant validity for the AI Leadership subdimensions relative to those measures. Criterion analyses further showed that AI Leadership scores added explanatory value across several work outcomes, with particularly large increments for AI adoption and integration in the 51-item sample. In the cross-validation analyses, content-family composites improved on the orbiting-construct baseline more frequently than subdimension blocks, indicating the value of examining both broader summaries and detailed behavioral scores.

## Theoretical Contribution

The framework organizes leadership behaviors around three recurring demands of AI-enabled work (Table 1), with the content families and behavioral subdimensions shown in Figure

1. Uncertainty in AI output and changing knowledge require leaders to update their understanding, make decisions under uncertainty, and seek knowledge beyond their teams. Changes in tasks and roles require transparency, skill development, team learning, psychological safety, and meaningful work. Accountability for AI implementation requires leaders to calibrate reliance on AI to the context, uphold ethical standards, and coordinate implementation over time. Bringing these demands together provides a way to examine how leaders coordinate their own judgment, the work of their teams, and the governance of AI use. It also supports research on relationships among these behaviors: how learning informs calibration, how disclosure supports shared interpretation, and how ethical boundaries shape implementation decisions.

The Battery focuses on leaders' actions within AI-enabled work. Research on algorithmic leadership and management examines how algorithms participate in directing and organizing work (Chang et al., 2026; Parent-Rocheleau et al., 2024). The present framework brings attention to how leaders interpret those systems, explain their use, organize the work around them, and accept responsibility for decisions. McGuire and De Cremer's (2023) finding that participants preferred human authority in a moral-leadership experiment illustrates why the source of authority matters in some AI-related decisions. The Battery provides behavioral measures for investigating how leaders exercise that authority through practices such as questioning AI assumptions, explaining AI use, and enforcing ethical boundaries.

**Measurement Contribution**

The Battery represents each of its 36 behavioral subdimensions with three items, giving researchers a consistent level of detail across the 11 content families. This structure makes it possible to formulate questions about specific practices, such as revising AI-use decisions, encouraging AI-related dissent, or explaining the rationale for AI use. The content families organize these practices theoretically, while the subdimension measures retain the detail needed to examine their different relationships with other constructs and outcomes.

The criterion analyses also examined equally weighted content-family composites as descriptive summaries. Comparing these composites with subdimension blocks allowed the research to evaluate broader and more detailed representations of the same behavioral content. Their different patterns of held-out performance (Table 8, Panel B; Figure 3) show why measurement detail should be considered in relation to the research question. The Inner, Outward, and Grounded Signals organize the content families for practitioner communication. The statistical analyses evaluated subdimension scores and content-family composites.

**Limitations and Future Research**

Future research can extend the Battery's initial measurement evidence by administering measures from all content families to the same participants, further evaluating the mixed model fit of the 51-item form, and examining discriminant validity among closely related subdimensions. Administering the measures together would allow researchers to evaluate relationships across the full Battery. Comparisons of alternative factor structures could clarify the level of measurement most useful for different research questions. Future research can extend the orbiting-construct comparisons by administering the full published scales.

Longitudinal studies and data from leaders, team members, and other observers can extend the present cross-sectional, leader-reported findings. These designs could examine changes in AI leadership practices and their associations with subsequent AI adoption, team learning, and implementation outcomes. Evaluating predictive performance in new samples would extend the cross-validation results, while comparisons across outcome measures could clarify which practices are most consistently associated with particular aspects of AI-enabled work.

Research across occupations, countries, languages, and levels of AI use would extend the current evidence from English-speaking leaders recruited through Prolific. Measurement invariance testing could evaluate whether the measures function comparably across groups and settings. Repeated measurement could assess test–retest reliability and, in leadership-development studies, sensitivity to change. Together, these studies would build on the Battery's behavioral framework to explain when, how, and for whom specific AI leadership practices matter.

## Conclusion

The AI Leadership Battery translates the demands of AI-enabled work into specific leadership behaviors that can be examined empirically. Its 36 subdimensions, organized within 11 content families, describe how leaders evaluate AI-supported judgment, organize learning and adaptation, communicate AI involvement, and maintain accountability during implementation. The research program combines content validation, exploratory factor analysis and item selection, confirmatory factor analysis in independent samples, and analyses of concurrent and incremental validity. These studies provide initial evidence for the Battery's use in investigating leadership practices and their relationships with organizational and AI-related outcomes. The Battery provides a common set of behavioral measures for research on how leaders guide work as AI becomes embedded in organizational activity.

## Ethics Oversight and Protected Materials

Elon University Institutional Review Board verified protocol 26-3582, "Artificial Intelligence (AI) Leadership: Developing a Measure of AI Leadership," as exempt on Jun 16, 2026.